\documentclass[a4paper, 11pt]{article}

\usepackage{a4wide}
\usepackage{amsmath}
\usepackage{amssymb}
\usepackage{commath}
\usepackage{latexsym}
\usepackage{graphicx}
\usepackage{color}
\usepackage[pdfstartview=FitH]{hyperref}
\usepackage{verbatim}
\usepackage{authblk}
\usepackage[subtle]{savetrees}

\title{Trustworthy Artificial Intelligence in the Context of Metrology\footnote{© 2024. This manuscript version is made available under the CC-BY 4.0 license \url{https://creativecommons.org/licenses/by/4.0/}. 
This is a preprint of the following chapter: T. Adel, S. Bilson, M. Levene and A. Thompson, \emph{Trustworthy Artificial Intelligence in the Context of Metrology}, published in \emph{Producing Artificial Intelligent Systems: The roles of Benchmarking, Standardisation and Certification}, edited by Maria Isabel Aldinhas Ferreira, 2024, Springer, reproduced with permission of Springer Nature. The final authenticated version is available online at: \url{https://link.springer.com/book/10.1007/978-3-031-55817-7}.  
}}

\author{Tameem Adel, Sam Bilson, Mark Levene and Andrew Thompson}
\affil{Department of Data Science, \\ National Physical Laboratory (NPL), \\ Hampton Road, Teddington, TW11 0LW, U.K. \\ \{tameem.adel, sam.bilson, mark.levene, andrew.thompson\}@npl.co.uk}

\date{\today}
\begin{document}

\maketitle

\begin{abstract}

We review research at the National Physical Laboratory (NPL) in the area of trustworthy artificial intelligence (TAI), and more specifically trustworthy machine learning (TML), in the context of metrology, the science of measurement. We describe three broad themes of TAI: technical, socio-technical and social, which play key roles in ensuring that the developed models are trustworthy and can be relied upon to make responsible decisions. From a metrology perspective we emphasise uncertainty quantification (UQ), and its importance within the framework of TAI to enhance transparency and trust in the outputs of AI systems. We then discuss three research areas within TAI that we are working on at NPL, and examine the certification of AI systems in terms of adherence to the characteristics of TAI.

\end{abstract}


\section{Introduction}

As background to the main story it is important to understand the meaning of {\em artificial intelligence} (AI), and more specifically how its subset {\em machine learning} (ML) fits into the picture. AI can be generally defined as the theory and development of computer systems that are able to perform tasks that normally require human intelligence. As such AI systems may be adept in discovering new information, making inferences and possessing reasoning capability. ML is a subset of AI focussing on AI methods that are able to learn and adapt. AI includes symbolic computation, such as expert systems, which are not a part of ML, whereas ML builds statistical models of data that may be used for classification and prediction tasks to aid decision-making. Here we focus on ML rather than AI, but will still use the term AI when referring to the more general technology.

\smallskip

An important distinction between an ML algorithm and a conventional algorithm (such as one that sorts a list of numbers) is that rather than solve the task directly, an ML algorithm produces a model, which is then used to solve the task at hand; the task could be a classification or regression problem. In classification the data is assumed to be divided into several {\em classes} and the task is to predict the class from the input data, while in regression the task is to predict a numeric value which is as close as possible to the true value. In order to generate the model, ML algorithms such as {\em neural networks} (NNs) process a dataset, which constitutes its input. In particular, deep learning \cite{GOOD17}, where an NN may have many layers, has been very successful in scaling applications beyond what was possible beforehand.

\smallskip

The word trustworthy means ``able to be relied upon as honest, responsible and truthful", which leads us to the concept of {\em trustworthy AI} (TAI) \cite{KAUR22,LI23}, or more specifically {\em trustworthy ML} (TML) \cite{VARS22}. As AI and ML applications proliferate, there is an increasing demand for demonstrating trustworthiness so that we can have confidence in the technology to make responsible decisions. In this sense TAI can be viewed as a framework for managing risk of potential negative impacts of AI systems \cite{TABA23}.

\smallskip

The principles of TAI can be broken down into three broad themes:
\renewcommand{\labelenumi}{(\arabic{enumi})}
\begin{enumerate}
\item {\em Technical} characteristics such as reliability, robustness, generalisability, resilience and security;

\item {\em Socio-technical} characteristics such as interpretability, explainability, bias freeness and privacy; and

\item {\em Social} characteristics such as transparency, accountability and fairness.
\end{enumerate}

It is evident that TAI must also address the social issues, which are of concern to society as a whole. These include ethical issues, which are of prime importance, although we will not elaborate on these any further. The socio-technical characteristics span the technical and social emphasising that TAI introduces a third dimension, in addition to the algorithms and models, which includes the interaction of humans with the technology. As an example, explainability, that is the ability to provide a clear explanation of why the AI system arrives at its outputs, is a key feature for promoting trustworthiness. Explainability also promotes fairness by increasing the transparency of the model, helping humans understand the decisions made by an AI system. It also addresses the 
``black-box" syndrome when a model is too complex to understand; this is especially prevalent in deep learning models, which typically have large numbers of parameters and features and may be hard to interpret.

\smallskip

The research we undertake at NPL, the UK's national metrology institute, in the TAI area (or more specifically the TML area) is done in the context of metrology, the science of measurement \cite{bipm2008,CROW20}. The International Vocabulary of Metrology (VIM) \cite{JCGM12} provides definitions for the vocabulary of metrology, which we would like to extend to encompass machine learning, where the measurement function could be, for example, a neural network. An NN could, in principle, be written as a collection of equations, and although it is not a physical measurement device, we should be able to trace back its inputs to some measurement device. From a metrology perspective, given an input dataset, an ML model generated by an ML algorithm, such as an NN, plays the role of a measurement model \cite{JCGM12}. However, as opposed to a traditional measurement model in metrology, where a physics-based equation is constructed according to the application, an ML measurement model will, generally, not have a straightforward physical interpretation. In fact, it is often treated as a black-box model, as its internal structure is not easy to understand in the context of the application at hand. Nonetheless, due to the complexity of real-world applications and the difficulty of producing precise physics-based models, both researchers and practitioners are turning to ML algorithms, which are data-driven, and with recent advances can operate on large 
datasets \cite{SEJN19}.

\smallskip

The VIM describes uncertainty of measurement as a parameter characterising the dispersion of values attributed to a measured quantity \cite{JCGM12}. The key point is that the parameter is derived from a probability distribution, which describes both the measured value and its uncertainty. It cannot be over-stressed that a measurement reported without an associated uncertainty is incomplete. Thus the quantification of model variable uncertainties and their propagation through the model are of central importance in metrology. Moreover, there is uncertainty both in the input data to the system (known as {\em aleatoric} uncertainty) and in the statistical model of the system that is used to make predictions (known as {\em epistemic} uncertainty) \cite{HULL21}. One can justifiably argue that the epistemic uncertainty present in ML-based models is, often, larger than that present in physics-based models.

\smallskip

{\em Uncertainty quantification} (UQ) can also be viewed as a characteristic of TAI that enhances transparency by not merely delivering a point estimate, which is not fully accurate due to the present uncertainties that we have just mentioned. As an example, a medical practitioner cannot make an informed decision regarding a test result without taking into account the uncertainties, which could be communicated via a margin of error, that is a confidence interval. General requirements that methods for the evaluation of uncertainty in ML for metrology should satisfy, were proposed in~\cite{thompson2021uncertainty} and are summarised in Section~\ref{subsec:app2}.

\smallskip

NPL has played a key role in setting the agenda for ML and AI as part of the Strategic Research Agenda (SRA) for the European Metrology Network for Mathematics and Statistics (EMN Mathmet) \cite{MATH23}. The SRA details mathematical and statistical issues that contribute to TAI and in particular to the trustworthiness of an ML prediction in the context of metrology. It also emphasises practical aspects of ML, including the importance of a quality framework for guiding the choice of an ML model, for supporting verification and validation of the ML algorithms and software used, and for considering the quality and provenance of the data.  The framework also supports the reproducibility of results and auditability of ML models in metrology applications. In addition, it highlights the importance of the specification of a standard interface for benchmarking, validation and certification of ML models. Finally, it presents a roadmap for the metrology community to meet the challenges of ML and AI.

\smallskip

Application areas of AI and ML at NPL are varied and include advanced manufacturing, energy systems, climate and earth observation and materials characterisation. We also work on various applications in healthcare including digital pathology for cancer research, and medical imaging for disease detection.

\smallskip

\begin{figure}[!ht]
    \centering
    \includegraphics[width=0.6\textwidth]{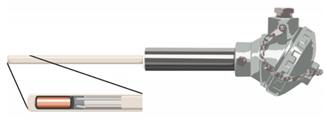}
    \caption[Self-validating thermocouple]{Self-validating thermocouple with protective sheath. Image courtesy of CCPI Europe.}
	\label{fig:thermocouple}
\end{figure}

To briefly describe a more specific example, NPL have developed an automated approach to calibrating thermocouples using ML~\cite{bilson2023machine}. Thermocouples measure temperature indirectly through a dependent property such as voltage. The voltage is related to temperature by comparison with a set of known temperatures, i.e., a calibration. The global framework for realising the SI unit of temperature, the Kelvin, is the International Temperature Scale of 1990 (ITS-90)~\cite{preston-thomas_international_1990}. A key drawback to this approach is that thermocouples can degrade when exposed to high temperatures, causing the indicated temperature to change in an unknown way. This degradation is referred to as ``calibration drift'', and is a major issue for industry applications where temperature monitoring and control is critical, such as nuclear waste storage and aerospace heat treatment. To address this issue, NPL's Temperature \& Humidity group have developed a self-validating thermocouple (see Figure~\ref{fig:thermocouple}), which is able to determine the level of calibration drift by identifying the time at which a metal ingot within the thermocouple melts, and then use the known melting temperature of the metal to re-calibrate the device. The device has been extensively characterised~\cite{tucker_integrated_2018}, and has been licensed by NPL to UK thermocouple manufacturer CCPI Europe, under the trade name INSEVA~\cite{wetton_ccpi_2018}.
\smallskip

To identify the time at which the metal ingot melts (the onset of melting), one must be able to identify a melting plateau in the indicated temperature from the self-validating thermocouple. Up until recently, locating and identifying the onset of melting has been performed manually through visual inspection. This is time consuming, and prone to human error. Thus, the ML group at NPL have devised a fully autonomous solution using a novel ML algorithm to first identify the onset of melting, and then provide uncertainty estimates in the ML model outputs. The ML model has a small number of interpretable parameters, defining the characteristics of the melting plateau. This will not only reduce time costs to meet industry needs, but also provides a trustworthy ML model through interpretability and UQ, to allow adoption in safety critical environments.

\smallskip

The rest of the paper is organised as follows. In Section~\ref{sec:tai} we discuss three research areas within TAI that we are working on at NPL. In particular, in Section~\ref{subsec:app1} we discuss the importance of explainable AI with examples of its application in metrological applications. In 
Section~\ref{subsec:app2} we describe a metrology framework for uncertainty quantification for ML. 
In Section~\ref{subsec:app3} we discuss good practice in training set preparation for object detection and classification systems in the context of marine autonomous navigation. 
In Section~\ref{sec:cert} we briefly discuss the issues in certification of AI systems in the context of TAI. Finally, in Section~\ref{sec:conc}, we present our concluding remarks.

\section{Research on TAI at NPL}
\label{sec:tai}

\subsection{Explainable AI}
\label{subsec:app1}

Given the recent advances in AI, it is becoming obligatory to understand the predictions obtained by AI systems. The quest for such an understanding represents one of the fundamental ethical issues calling for consideration in relation to AI. In addition to explainability, other issues include fairness and accountability. The main driver for addressing these ethical issues is the need to mitigate the negative impacts of AI on society. Explainable AI therefore represents one of the main cornerstones in the quest to promote trust in AI systems. 

\subsubsection{Motivation} \label{section:explnbl_motiv}

AI systems are nowadays applied to a wide scope of areas, including sensitive applications such as self-driving cars \cite{MA20} and medical diagnosis \cite{KUMA23}. Therefore, it is not acceptable anymore to blindly trust black-box AI systems solely based on the fact that they are capable of reaching high levels of accuracy. This is one of the main reasons why there has recently been a growing interest in developing explainable AI models and systems \cite{Adel2018,Antoran2021a,Poursabzi-Sangdeh2021a}. 

\smallskip

The term black-box, mentioned in the introduction, refers to a system or a model whose inner reasoning is completely hidden to its user. As far as ML is concerned, this typically refers to a model which, given a particular input, can output a certain prediction via a process which is not understandable to the external user.

\smallskip

The need for automating systems via AI has a strong foundation \cite{RUSS21}. Furthermore, AI systems have the capacity to be faster, more accurate, and more consistent than their human counterparts. However, there are exceptions. Explainable AI aims to provide the system user with the opportunity to better understand the challenges an AI system can encounter such that these exceptions can be addressed before integrating the system, especially in applications whose nature enforces a tiny tolerance for errors. In addition, the ability to understand AI models and systems is also related to fairness, since people would be more tolerant about how unfair a certain model is if the model can explain the reasoning behind its own predictions. 

\smallskip

NNs, for example, can also behave unexpectedly in certain situations. For example, imperceptible alterations (such as those performed when designing adversarial examples) to image pixels can lead to a change in the prediction of the corresponding class \cite{Nguyen2015a,Gilpin2018a}. The potential for unexpected behaviour of NNs can be mitigated by making them more explainable. 

\smallskip

Explainable AI supports metrological traceability. The latter addresses measurement systems which make use of AI and links the measurements with a standard as a point of reference, which implicitly implies a consistent level of understanding among the system users. Correspondingly, explainable AI also aims to provide reference standards in order to render the overall predictions and decisions obtained by an AI system understandable to users.

\subsubsection{Neural network explanations} \label{section:explnbl_cncpts}

Regarding the explanations of NNs, one taxonomy \cite{ARRI20} divides them into two categories. The first category is focussed on explaining the reasoning behind the input/output pairs. In other words, the focus of this category is to answer the following question: given a particular input, why does the NN provide us with the corresponding output? The second category aims to concentrate on the internal representation of the NN. 

\smallskip

The first category, which aims at explaining the predictions related to input/output pairs, is currently the most popular category. Recent advances in the design of neural network architectures have led to a dramatic increase in network sizes. Nowadays, a network containing half a billion ($5 \times 10^8$) parameters is considered an average-sized NN. The bottleneck here is to explain the manner by which the respective NN has processed the input all the way through until the produced output, while ensuring that the explanation is considerably less complex than the original NN. One way to achieve this is via creating a salience map, which highlights the most relevant parts of the input \cite{Zintgraf2017a}. In other words, the salience map would tell us which parts of the input are the most responsible for the obtained prediction. Another way is to create a proxy model, which is simpler than the original model, yet its behaviour is as similar as possible to it. 

\smallskip

A prime example of the proxy model approach can be found in \cite{Ribeiro2016a}. The method therein is referred to as {\em local interpretable model-agnostic explanations} (LIME). The main idea of LIME is to explain a black-box NN by first perturbing the input, then using the perturbations to establish a local linear model. This then acts as a proxy which is simpler than the original black-box NN, yet it explains the NN within the neighbourhood of the input. Provided the original image input is divided into superpixels, LIME can, as well, be used to identify input regions which are the most relevant for the NN predictions; superpixels are regions which have a meaning that can be understood by the model experts. The fact that superpixels must be identified a priori makes LIME more difficult to deploy for some applications where such expert knowledge is not available. 

\smallskip

The {\em prediction difference analysis} (PDA) algorithm developed in \cite{Zintgraf2017a} represents another seminal explainability method, which is based on developing salience maps. The PDA method visualises the response of an NN to a specific input. For an NN used to classify images, PDA highlights the areas in a given input image which provide evidence for or against a certain class. PDA overcomes several shortcomings of previous methods, including LIME, since it does not require specifying superpixels a priori, and is basing the reasoning followed to identify the pixel relevance values on a principled Bayesian procedure. PDA has also been applied to the healthcare domain and MRI brain scans \cite{Zintgraf2017a}. 

\smallskip

Other algorithms aiming to explain the decisions of an NN by producing salience maps include the work in \cite{Zeiler2014a}, which tests the network via occluding parts of the input in order to construct a map demonstrating which input parts are more influential with respect to the network output. 

\smallskip

Quantifying explanatory features has also been applied to metrology. For instance, the method in \cite{Sadowski2015a} combined the self-explanatory {\em principal component analysis} (PCA) with a self-organisation feature map in order to determine the pull-off adhesion between concrete layers in floors.

\smallskip

Another,  more classic, approach to explain predictions of NNs is based on automatic rule extraction. For example, the work in \cite{Bologna2019a} simplifies the decisions of a {\em convolutional neural network} (CNN) by performing rule extraction on its fully connected layer, with the help of {\em discretised multi-layer perceptrons} (DIMLPs). Other approaches to extract rules aim to find trends from the input to the output, like the validity interval analysis method in \cite{Thrun1995a}, which applies sensitivity analysis to mimic the NN network behaviour. The method in \cite{Hailesilassie2016a} makes use of a genetic algorithm to generate new training examples, whereas the work in \cite{Taha1999a} applies a sampling approach to produce (binary) input data points. 

\smallskip

Examples of the second category, which focus its explanation on the network's internal representation, include \cite{Simonyan2013a} where visualisations of the input patterns maximising the response of a single unit are created using gradient descent, for the sake of understanding the role of each individual unit. Similar methods have been applied in the metrology field. For example, the work presented in \cite{Li1995a} aims to quantify the role of the different parts of the NN, which was developed for automatic arc welding purposes. 

\smallskip

Methods which focus on explaining the network representation usually characterise the role of every individual hidden unit by testing the whole network representation on a similar, yet not identical, task in a transfer learning setting. Hidden units are then measured according to their ability to detect specific understandable concepts across the transfer \cite{Gilpin2018a}. The same strategy can as well be applied to network layers, rather than individual hidden units, especially in cases where it is believed that all the units belonging to the same layer share a similar behaviour.

\smallskip

The work in \cite{Niri2022a} depicts another application of explainable AI in metrology, where the method exploited two explainable ML indices to systematically analyse the impact of slurry coating on the manufacturing process of Li-ion battery electrodes.

\subsection{A metrology framework for uncertainty evaluation in machine learning}
\label{subsec:app2}

\subsubsection{Motivation}

While the most celebrated recent advances in ML have been upon classification tasks, a typical use of ML in metrology is in a regression task in which a continuous quantity is to be estimated. Although the goal is to use information about a continuous quantity to make decisions or to assess whether a system meets specific requirements, the fundamental challenge is to measure the quantity to provide that information, and so we focus in this article on regression problems. 

\smallskip

How do regression problems arise in metrology?  They arise because it is often not possible or desirable to measure a quantity directly, but rather it is necessary or desirable to infer information about a quantity from a number of contributions which are easier to measure or for which information is available. These contributions could relate to measurements made by some instrument or system, to applied corrections, or to information obtained from sources such as manufacturer's specifications or calibration certificates. A model is then needed to infer information about the quantity of interest from these contributions. The framework of evaluating uncertainties by means of measurement models was standardised for the metrology community in the influential ``Guide to the Expression of Uncertainty in Measurement'' (GUM)~\cite{bipm2008}.

\smallskip

There are many scenarios in which it is not possible to build or perform computations on an analytical model based upon physical understanding, and instead a data-driven approach must be taken. Take three example areas of measurement science at NPL, where ML is either currently used or where it offers potential.
\begin{itemize}
    \item In environmental monitoring, the data provided by sensors is used for forecasting to understand how the state of the environment is changing with time and the impacts of different sources, including anthropogenic sources, on the environment. An example is the use of underwater {\em hydroacoustic sensors} (hydrophones) to provide data about levels of sound in the oceans to understand how human activity is affecting those levels. For many applications, the physical processes are too complicated to be analysed theoretically. In these cases, methods based on ML, including Gaussian processes and recurrent neural networks, provide a viable solution~\cite{robinson2023impact}.
    \item {\em Mass spectrometry} (MS) is a suite of surface analysis techniques that measure the chemical composition of a sample. This is achieved through removal of material from the surface (e.g desorption) and recording the intensity of a particular mass to charge (m/z) ratio of the removed material. The resulting data are high dimensional and heterogeneous intensity spectra containing peaks that characterise particular compounds of the analysed material~\cite{thomas2016dimensionality}. The spectra obtained are contingent on the physico-chemical properties of the sample, process of material removal, and the instrumentation measuring the intensities, hence analysis is difficult. 
    \item In earth observation and remote sensing, inter-instrument bias is a ubiquitous problem, but a complete theoretical understanding of the causes of such biases is rarely available. ML has therefore been used for bias correction and cross-calibration of sensors in the context of various types of measurement including vegetation indices and ozone depletion. ML has also been used to build data-driven models which infer measurements of airborne particulates and pollen estimation. We refer the interested reader to the review article~\cite{lary2018machine} for more details.
\end{itemize}

The solution, then, in metrology applications such as these, is to learn a measurement model from calibration data. 

\smallskip

Suppose that we are interested in measuring some quantity $Y$ indirectly; we will refer to this quantity as the \emph{measurand} or the output quantity. We have access to a set of \emph{training data} comprising multiple values of the measurand $Y$ and corresponding values of the contributions $X_1,\ldots,X_N$ in the measurement model.

\smallskip

The first stage of the regression problem is to use the training data to learn an approximate measurement model $Y\approx f(X_1,\ldots,X_N)$. The typical approach in ML is to restrict $f$ to some parametrised family of models and then learn its parameters. 
The second stage, having learned a model $f$, is to use it to obtain an estimate of the measurand $Y$ along with an assessment of its uncertainty, given similar information for the input quantities $X_1,\ldots,X_N$. 

\smallskip

One of the foundational principles of uncertainty evaluation in metrology, systematised by the GUM framework~\cite{bipm2008}, is that uncertainty is not an intrinsic property of a measurand, but rather a \emph{probabilistic statement of belief} based upon our perceived knowledge. For example, uncertainty is defined in the GUM as ``a parameter associated with the result of a measurement, that characterises the dispersion of the values that could reasonably be attributed to the measurand''. The most comprehensive form that such a statement might take is the {\em probability density function} (pdf) of the measurand $Y$. In an ML context, only a numerical approximation to the pdf is realistic, which might be captured by an evaluation of certain quantiles of the distribution. 

When it is legitimate to make assumptions concerning the distribution, for example that it is a Gaussian, information might be captured by summary information such as the mean (often referred to as the \emph{best estimate} $y$) and standard deviation (often referred to as the \emph{standard uncertainty} $u(y)$) of $Y$. Coverage intervals which contain $Y$ with a stated probability $p$ might also be obtained. Symmetric coverage intervals of the form $[y-U_p,y+U_p]$, where $U_p=k_p u(y)$ is an \emph{expanded uncertainty} and where $k_p$ is a \emph{coverage factor}, might be obtained. Alternatively, a coverage interval of shortest length for a given probability $p$ might be chosen (the two approaches are equivalent if the distribution is unimodal and symmetric about the mean). 
In addition, it is realistic to expect that similar statements of belief about each of the input quantities $X_1,\ldots,X_N$ are available.

\smallskip

The importance of the GUM to the metrology community stems from how it aims to provide harmonisation and transparency in the way uncertainties are evaluated and reported, and one way it does this is to try to bring clarity to the problem intended to be solved when evaluating an uncertainty. For example, in the case that the measurement model is known and the pdfs for the input quantities are fully specified, the pdf for the measurand is completely defined and the challenge is to determine that pdf to a degree of approximation that is adequate for how the pdf will subsequently be used. In the context of ML, an equivalent level of harmonisation and transparency would seem to be lacking.

\smallskip

The GUM framework allows for different approaches to determining the pdf for Y, including the algebraic determination of the pdf when that is possible, appealing to a combination of the \emph{law of propagation of uncertainty} and the central limit theorem, and numerical approaches such as a Monte Carlo method. Excepting the first of these approaches, the solutions obtained are generally approximate, and so there remains the aspiration to understand the degree of approximation when these approaches are applied in specific cases. Since uncertainty is conceived of as an expression of belief, a Bayesian framework is considered by many to be preferable in metrology, and there have been proposals to revise the GUM entirely along Bayesian lines~\cite{bich2012revision}.

\smallskip

Conventional uncertainty evaluation of the kind systematised by the GUM framework makes the assumption that the measurement model is known. Uncertainty evaluation in ML deviates from this paradigm in one key respect. An ML model is not deduced from physical understanding, but is learned from data. It follows that, even after an ML model has been learned from data, the absence of physical insight leads to uncertainty in the model itself. One way to address this challenge is the use of physics-informed ML~\cite{karniadakis2021physics}.

\smallskip

More recent work at NPL has focused on the use of Bayesian inference methods for uncertainty evaluation in parametrised models. These approaches learn by combining data with prior knowledge, and replace the measurement model with a model of the observation made by a measuring system or instrument~\cite{forbes2011gum,klauenberg2015tutorial}.  Learning measurement models in ML is closely related conceptually, but the main difference is that in an ML context little can be assumed about the form of the model, and heavily overparametrised models must be learned based on large volumes of data.

\subsubsection{Metrology requirements for uncertainty evaluation}

\begin{figure}
\centering
\includegraphics[width=0.45\paperwidth]{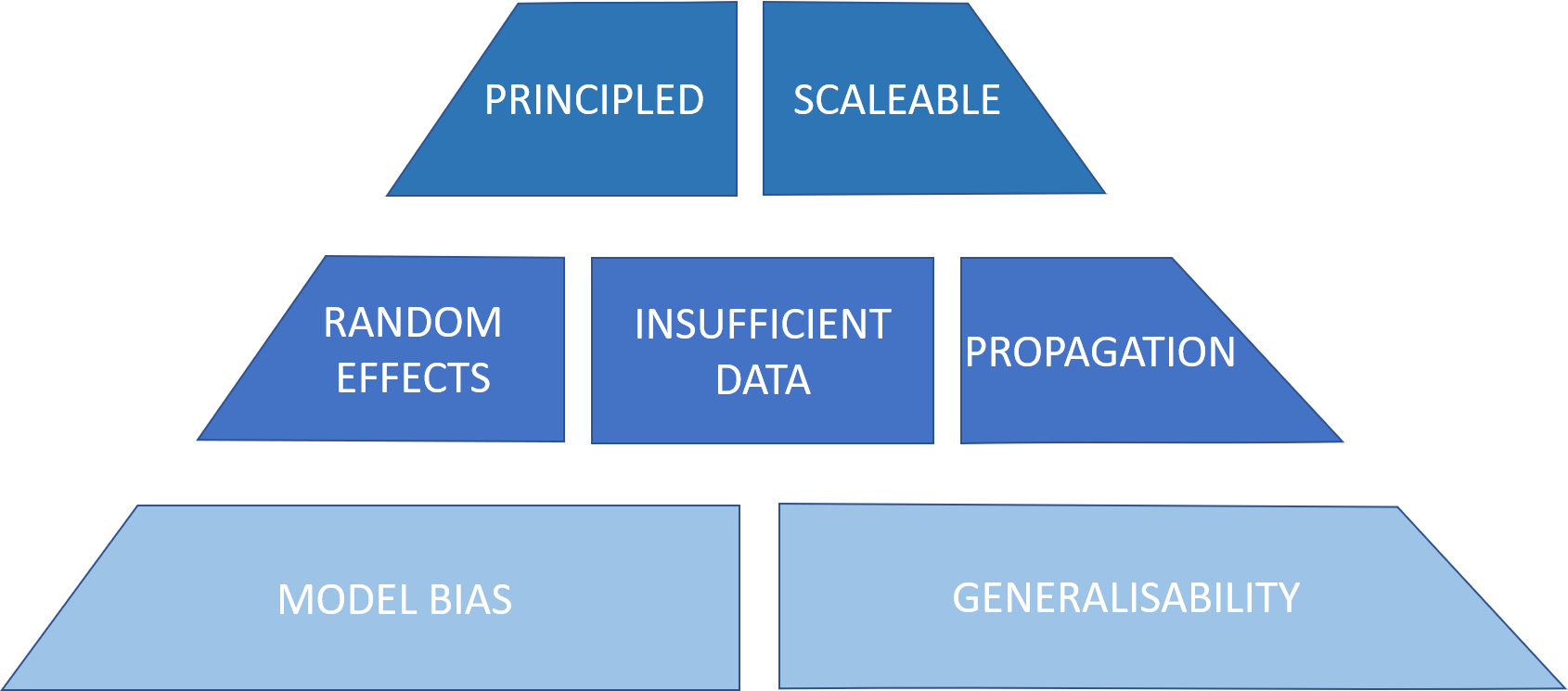}
\caption{Metrology requirements for machine learning uncertainty evaluation.}\label{requirements}
\end{figure}

In~\cite{thompson2021uncertainty}, we presented a metrology perspective on the requirements that uncertainty evaluation methods for ML should meet. We identified seven requirements, which are illustrated in the form of a pyramid in Figure~\ref{requirements}.

\smallskip

\paragraph{Model bias.} The two requirements at the base of the pyramid arise from a simple observation: \emph{no uncertainty evaluation of an ML model can be trusted if the ML model itself cannot be trusted}. One way to express the two aspects of this requirement is that every effort should be made that the ML model must neither underfit nor overfit the training data.

\smallskip

If a family of models is chosen which is not sufficiently expressive to capture the relationships between the input and output quantities, underfitting will occur. A simple example would be performing linear regression when the relationship between the data is nonlinear. 

\smallskip

Another way to describe this effect is as one of unwanted model bias. There is an inescapable tradeoff between bias and variance in ML models, and fitting a model that in any way generalises is only possible by introducing some degree of bias, either by restricting to a certain model class or through regularisation techniques in learning the parameters of the model. However, for uncertainty evaluation there is an imperative to eliminate model bias as far as possible. Underfitting leads to inaccurate uncertainty evaluation because, if relationships in the data have not been fully discovered, the model is ignorant about what it does not know. 

\smallskip

This principle can be traced in the GUM framework, where care is taken to distinguish between random and systematic effects contributing to measurement error, and where, ``It is assumed that the result of a measurement has been corrected for all recognised significant
systematic effects and that every effort has been made to identify such effects''~\cite[Section 3.2.4]{bipm2008}. 

\smallskip

\paragraph{Generalisability.} The requirement for generalisability is the flipside to the previous requirement. While it is important that a model is sufficiently expressive to capture the systematic relationships between the measurement data, it is important that the model does not overfit the training data in a way which does not generalise. The danger of overfitting is well known in ML, and it is also well known that it is a particular danger for very expressive model classes such as {\em deep neural networks} (DNNs). 

\smallskip

\paragraph{Random effects.} The middle layer of the pyramid focuses upon the different sources of uncertainty that need to be captured. Applying an ML model can be viewed as a two-stage process: a training phase followed by a prediction phase. In the training phase, a measurement model is learned from training data, and then in the prediction phase this measurement model is applied to new input data. Uncertainties are introduced in both stages of this process. During the training phase, uncertainty concerning the model itself is introduced. Model uncertainty arises due to either insufficient coverage of or uncertainty in the values of training data. 

\smallskip

Having learned the model, uncertainty also arises in evaluating the model on new input data. In addition to the uncertainty concerning the model's parameters, uncertainty in the input quantities is also propagated through the model. Provided systematic effects have been removed, this uncertainty can be equated with random effects. 

\smallskip

The two main sources of uncertainty identified here correspond to a distinction that is often made in recent literature between \emph{aleatoric} and \emph{epistemic} uncertainty~\cite{HULL21}. Aleatoric uncertainty refers to uncertainty which is intrinsic to the problem, and by which is usually meant random effects. Epistemic uncertainty, on the other hand, refers to uncertainty concerning the model. Capturing both types of uncertainty is crucial from a metrology standpoint.

\smallskip

Random effects must be taken into account, because otherwise incorrect assumptions are being made about the model, namely that there exists a correct model, uncorrupted by random fluctuations, thereby invalidating the uncertainty evaluation carried out upon the model. The requirement here, then, is precisely that aleatoric uncertainty be properly captured. 

\smallskip

\paragraph{Insufficient data.} It is widely acknowledged that it is dangerous to blindly extrapolate an ML model to regions of data space where no data has been seen. Intuitively, given some input data, our uncertainty in the output of a model should be related to the extent to which we have observed similar input data in training. It follows that, if epistemic uncertainty is not captured and uncertainty evaluation is restricted to estimating noise distributions, our uncertainty evaluations will be overly optimistic and will not capture uncertainty due to insufficient data.
\smallskip

\paragraph{Propagation.} We recall that central to the GUM framework is the concept of a measurement model, and that uncertainties are evaluated by propagating input uncertainties through the model. The approach to propagation espoused by the GUM is the so-called \emph{law of propagation of uncertainty}~\cite[5.1.2,5.2.2]{bipm2008}, which relies on Taylor approximations. It is valid provided the measurement model can be assumed to be locally linear (or locally low-order polynomial) and providing the output quantity can be assumed to be either Gaussian or $t$-distributed. An alternative Monte Carlo approach is recommended in Supplement 1 to the GUM~\cite{bipm2008supplement}, which is useful when the above assumptions are not valid.

\smallskip

Many current approaches to uncertainty evaluation in ML only explicitly assume random uncertainties in the output quantity and do not explicitly model random uncertainties in input quantities. The problem is typically posed as one of standard regression, whereas errors-in-variables regression \cite{GILL10} is commonly used to assess the impact of input uncertainty. 
In an ML context, the situation is complicated by the fact that the model itself must be learned from data. It follows that simply propagating input uncertainties through a fixed model ignores parameter uncertainty, and what is needed instead is to combine uncertainties in data with parameter uncertainty (which itself also partially arises from uncertainty in training data). 

\smallskip

\paragraph{Principled and scalable methods.} At the top of the pyramid are two high-level requirements which constitute the ultimate goal, namely that uncertainty evaluation is both \emph{principled} and \emph{scaleable}.

\smallskip

The focus of much recent research into uncertainty evaluation for ML has focused upon the challenge of finding approaches which are scaleable to large problems. For example, classical approaches to Bayesian inference when applied to DNNs  do not scale well, leading to a search for methods which evaluate uncertainty more cheaply.

\smallskip

However, such attempts to reduce computational expense usually mean that principles have to be sacrificed to some extent. We have observed that uncertainty has precise definitions in metrology, and that vague attempts to quantify uncertainty are unlikely to be acceptable in this context. It is therefore important that, in attempts to scale up uncertainty quantification, principles are maintained as much as is possible.

\subsection{Good practice in training set preparation for marine navigation systems}
\label{subsec:app3}

In this section we summarise a good practice framework for marine autonomous navigation systems developed at NPL which was originally presented in~\cite{khatry2021good}.

\subsubsection{Background}\label{background}

There is growing interest in the automation of marine vessels, and in particular navigation systems which use AI. Some of the components of an AI marine navigation system require the training of ML models from data. All ML models are only as good as the data used to train them, and so ensuring that good practice is followed in the preparation of such training data is an important concern. 

\smallskip

An autonomous navigation system can be viewed as a pipeline consisting of four main steps: {\em sense-understand-decide-act} (S-U-D-A). The first step is sensing, in which information is collected from sensors mounted on the mast of the ship, including cameras (images), radar (distance information), GPS/IMU (location/position) and AIS ({\em automatic identification system}, which provides information about other vessels). In the understand step, the images are used for object detection and classification, the results of which are then fused with data from other sensors to infer locations of all obstacles including where the ship is in that environment. This is also merged with the estimated trajectories of the moving obstacles to get the available and non-available movement regions. The decide step then uses this information in path search algorithms to calculate the optimum trajectory that the ship needs to take while considering the moving and static objects. In the act step, this information is then fed to the control system to create appropriate control commands which is passed to the actuation system to make appropriate movements of the vessel.

ML is used mainly in object detection and classification component of a marine autonomous navigation system. The object detection task involves identifying the presence of a vessel along with a bounding box giving its location, while the classification task involves additionally determining the type of vessel. It is important to note that the two tasks should be distinguished, and the system requirements for detection and classification may be different. For example, it may be required to detect remote vessels and to perform low-granularity classification sufficient to identify extremely large vessels (for example mega-containerships), while it may be required to perform a higher-granularity classification once the vessels are closer. Similarly, the granularity of the classes must also be informed by the user requirements (for example it may be sufficient to distinguish between ships and buoys, or it may be necessary to distinguish between different types or sizes of ship).

\smallskip

The datasets required to train such a machine learning algorithm must consist of the following characteristics:
\begin{enumerate}
    \item The domain images: the images which are taken in the operational design domain of the ship.
\item The bounding boxes: boxes to denote the ship within the image which is used for training the object detection system (as well as the output of the object detector).
\item Object segmentation: pixelwise location of the image (optional).
\end{enumerate}
\smallskip

There are two main types of approach to ship detection in a maritime environment. Classical computer vision-based methods break the problem down into a number of technical tasks such as horizon detection, background subtraction, foreground detection and tracking. While this approach maintains a high level of explainability, these methods are highly sensitive to changes in background and illumination conditions. 

\smallskip

As a result, newer deep learning techniques are now considered state of the art. These approaches can be thought of as specialised CNNs, which combine feature detection and image classification. Deep learning approaches are potentially less sensitive to variations in background and illumination, though it still needs to be ensured that the model training process ensures that it is agnostic to such conditions.

\smallskip

An example of a CNN architecture for object detection is the Single Shot MultiBox detector~\cite{liu2016ssd}. In this architecture, an image is passed through a 16-layer image classification network (VGG-16) before being passed through additional layers which predict the offsets to default bounding boxes of different scales and aspect ratios and their associated confidences. 

Another approach often used in conjunction with deep learning is transfer learning~\cite{yang2020transfer}, in which an existing model is adapted for a new dataset, rather than learning a new model from scratch. In this approach, the final layers of the CNN are typically stripped away to preserve the feature extraction capability, while the new classification layers are created and retrained on smaller volumes of new data. This accelerates the transfer of the pre-trained model onto a new domain (domain shift). 

\subsubsection{A good practice framework}

\begin{figure}
\centering
\includegraphics[width=0.45\columnwidth]{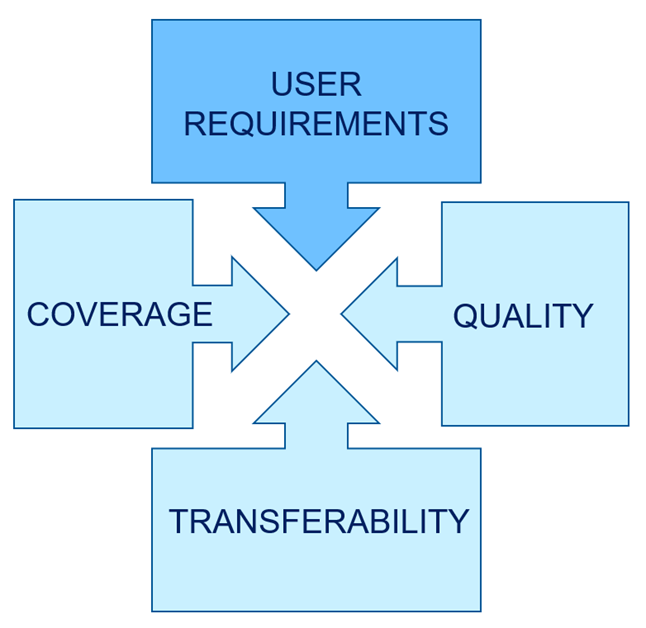}
\caption{Areas of good practice for training data preparation.}\label{training_framework}
\end{figure}

In this section, we highlight three areas of good practice which are important to consider when collecting and preparing an image dataset for object detection and classification. In addition, an important general observation is that the ultimate measure of good practice in this context is whether the training data leads to an ML model which meets the system requirements. The areas of good practice are shown in Figure~\ref{training_framework}.

\smallskip 

These areas of good practice align significantly with three of the popular four Vs of big data~\cite{patgiri2016big}. The fourth V of big data is \emph{velocity} (the speed of data acquisition), and this is less relevant in the context of training data. However, it certainly becomes important when it comes to deploying a marine navigation system where multiple data streams need to be evaluated within a narrow time window.
\begin{enumerate}
   \item \emph{Volume} (the number of examples in the dataset)
\item \emph{Variety} (the extent to which the dataset captures all possible scenarios)
\item \emph{Veracity} (the accuracy of the dataset).  
\end{enumerate}
\smallskip 

We next summarise what is meant by each area of good practice.

\smallskip 

\paragraph{Data and model transferability.} Object detection and identification involves learning a complex and high-dimensional classification model. Statistical learning theory tells us that, in such a setting, a large training dataset is required to train an accurate model. From this perspective, if there exists a dataset consisting of marine vessels, of a type relevant to the classification task at hand, it may make sense to use it. However, it is also important to be aware that an existing dataset may not be a good fit to the specific scenario of interest. For example, if two datasets were collected in the Antarctic and warm water regions respectively, one would expect the classes (types of vessels), lighting conditions and image quality covered by the two datasets to be different. Equally, image quality is likely to vary between two datasets which are not collected using the same sensors. Given such differences, a classification model built for one dataset may not be transferable to another dataset (a scenario often referred to as domain shift). If there is domain shift, it will be necessary to use transfer learning to retrain the model on new data as described in Section~\ref{background}.

\smallskip

In the fortunate situation that a model has already been trained on an existing dataset, then it may not be necessary to train a new model. However, the question of domain shift must be asked: is the model likely to be valid for the specific object detection task of interest? While domain shift might mean that a model is not entirely transferable, transfer learning as described in Section~\ref{background} can be used to adapt an existing model. This approach still requires a modest volume of new data to be available, but it will likely reduce the amount of data that needs to be collected.

\smallskip

\paragraph{Data coverage.} As was noted previously, object detection involves learning a complex and high-dimensional classification model, and in such a context the requirements on the training set are that there is sufficient data to be representative of the range of different object detection and classification tasks expected to be encountered by the marine vessel. It is crucial therefore that the specific functions that the object detection system will be required to perform are used to inform the decisions about data collection from the outset. In addition, the dataset must have adequate volume and variety~\cite{patgiri2016big}. Volume in this context simply means the number of training samples, whereas variety is about how well the dataset captures the variations between the classes and within the same class. 

\smallskip

In the context of training images for marine autonomy object detection, in order to sufficiently capture between-class variations, class imbalance should be avoided, in which the number of sample images of some types of vessels is much greater than others. When it comes to within-class variations, invariance to different viewing conditions (for example illumination, background and presence or otherwise of occlusions) is the main concern in a marine autonomy setting. 

\smallskip

Data augmentation~\cite{zoph2020learning} refers to the practice of artificially generating additional training data to enhance an existing dataset. Whether data augmentation should be performed will depend upon whether there is a need to correct for inadequacies in the collected data.  There are three main reasons why data augmentation might be required.

\begin{enumerate}
\item To improve the robustness of a machine learning model to intra-class variations (such as differences in viewing conditions).
\item To correct for class imbalance, namely that there are more data samples in some classes than others. 
\item To increase the size of a training set if the training set is originally too small.
\end{enumerate}
\smallskip

Simple examples of data augmentation include rotation of images or parts of images to simulate different viewing angles, and the use of image adjustment techniques to simulate different lighting conditions, geometric distortions, obstructions and viewing angles. Data augmentation should be subjected to the same level of scrutiny as the original data: it is important that the augmented data is representative of real-world conditions. For example, if rotations are used, the rotations should be representative of the viewpoints and environmental conditions that are expected to be encountered by the marine vessel during operation of the navigation system. 

\smallskip

\paragraph{Data accuracy.} Inaccuracies in training data lead to inaccurate machine learning models, and consequently uncertainties in training data propagate through a model to give uncertainties in the classifications. With respect to training images for object classification, there are two aspects of data accuracy (or veracity~\cite{patgiri2016big}) that must be addressed: labelling quality and image quality.

\smallskip

Concerning labelling quality, creating a labelled database of marine vessel images will need to be carried out manually, which can be a time-consuming process. Labelling can either be carried out in-house or by using crowd-sourcing services, e.g. reCAPTCHA. It is vital that labelling correctly captures ground truth, whilst at the same time identifying images, which are known to the labeller because of other references but not identifiable through object features. There exist formal recommendations on custom labelling of datasets, for example the {\em Visual Object Classes} (VOC) Annotation Guidelines~\cite{VOC2007}.

\smallskip

Labelling and annotation are manual processes and are inherently subject to human error. It is therefore important that confidence in the quality of the labelling and annotation is established by assessing the labelling and annotation quality. This can be achieved either by assessment against ground truth if it is available, or by assessing inter-observer variability~\cite{joskowicz2019inter}.

\smallskip

Concerning image quality, images have an inherent accuracy limit determined by their resolution. The resolution of an image is determined by two factors: detector resolution (the number of detector elements) and the captured image resolution (number of pixels). The issue of resolution comes into play especially when designing specifications for datasets for classifiers for detecting ships near the horizon. In addition, weather effects can reduce image quality by introducing noise, blur and distortion. The performance of an object detection and classification system would be expected to degrade as image quality degrades. 

\smallskip

Another aspect that needs to be kept in mind is that the precise way in which performance degrades with image quality will be highly dependent upon the detection/classification task. For example, in satellite imaging objects of 5 pixels in size can sometimes be localised with high confidence, whereas in face identification it has been reported that the object size needs to be 32 pixels to be identified with reasonable confidence~\cite{torralba2009many,mansour2019automated}. Hence it can be understood that acceptable resolution of an object is related to the kind of detection/classification task that needs to be performed. The same principle would apply to noise from weather conditions. In a marine autonomy context, it would be easier to distinguish between boats and islands but, in order to distinguish which kind of boat or boat model, higher quality images would be needed. Viewed another way, if the resolution is held fixed, distinguishing type of boat or boat model would only be possible at a relatively shorter distance compared to just detecting a boat etc.

\smallskip

\paragraph{User requirements.} It is vital to emphasise that the ultimate measure of good practice in this context is whether the training data leads to an ML model which meets the system requirements. We have noted in Section~\ref{background} that system requirements need to be determined carefully, and that there may be various modes of operation (for example far-field detection versus accurate classification of nearer vessels). Navigational safety regulations such as the International Regulations for Preventing Collisions at Sea (COLREGs)~\cite{COLREGs1972} provide operational requirements on the performance of autonomous marine navigation systems, and consequently upon its various components, including object detection and classification. It is vital therefore that empirical testing is carried out to assess the performance of the ML models against these system requirements. These empirical results will reveal whether the machine learning model meets the system requirements. 

\section{Certification of AI systems}
\label{sec:cert}

This section summarises the main points presented in \cite{LEVE23}. We first need to distinguish between standards, certification and regulation in the context of AI systems. {\em Standards} inform organisations how to manage their AI systems, and foster a common terminology that provide a basis for the design, engineering and use of these systems. {\em Certification} of AI systems, on the other hand, is a third-party service whose aim is to assess the conformity of an AI system to some set of criteria. While standards are not legally binding as such, and certification, is generally, optional, {\em regulations} are enforceable policies created by government or regulatory bodies.  In the context of AI and ML, standards and certification can be viewed as market-based tools that are used by regulators to address various aspects of TAI pertaining to the risk present in the development and usage of AI. Regulation of AI systems should: encourage transparency so that its users will be able to understand its outputs and make responsible decisions; hold the relevant stakeholders accountable for harms caused by the development, deployment and use AI systems; promote fairness in the sense that AI systems should be accessible to all parts of society without discrimination and their outputs should not be used for discrimination of any group of people in society; and respect privacy of users data in accordance with data protection laws.

\smallskip

In \cite{LEVE23} we argued that certification of an AI system should take place along three dimensions:
\renewcommand{\labelenumi}{(\roman{enumi})}
\begin{enumerate}
\item against standards such as ISO/IEC FDIS 42001 pertaining to the management of an AI system,
\item against criteria pertaining to the trustworthiness of the system, and/or
\item conformity testing and evaluation of the AI components of the system.
\end{enumerate}
\smallskip

It is paramount that the human factors, being an integral part of the socio-technical characteristics of trustworthy AI, are taken into account in testing and evaluation, as the interaction between the AI system and humans demands measures beyond the traditional ones \cite{FREE22}.

\smallskip

We briefly mention  {\em Large language models} (LLMs) and the more general foundation models, as they pose a host of problems with regards to certification and regulation. LLMs are generative neural networks, traditionally trained on very large text corpora and web data, that  have a very large number of parameters on the order of billions or more. {\em Foundation models}, \cite{BOMM22} generalise LLMs to multimodal data, although it is common to use these two terms synonymously. Examples of operational LLMs are, OpenAI ChatGPT ({\em Chat Generative Pre-trained Transformer}) \cite{CHAT23} and Google Bard \cite{GOOG23}.

\smallskip

An LLM can be interrogated directly through its conversational interface (that is the chat), and can also be adapted to provide a solution to a downstream task through transfer learning. A downstream task is a specific task requiring a solution, such as a classification task, and the transfer learning mechanism used is known as fine-tuning.
LLMs such as ChatGPT and Bard are a disruptive technology that may affect user interaction with search engines and potentially revolutionise automated question answering platforms. We leave further speculation in this direction to others.

\smallskip

We give several examples where LLMs provide serious challenges with regards to whether they are trustworthy when tested against the TAI characteristics. One example, where privacy is potentially violated, is when personal data was used in training an LLM. In this scenario the personal data is extracted from the LLM and then used in an internet search; this is known as an inversion attack. Another example is when the output of an LLM is interpreted by the user as a form of advice, which potentially may be misleading. Yet another example is potential infringement of copyright when an LLM does not cite the source of its answer. Furthermore, bias may exist in the data on which the LLM was trained on and remain present in its answers, also violating the principle of fairness.

\smallskip

The question is who is responsible for such violations of TAI by an LLM \cite{HACK23}. Is it (i) the {\em developers} of the LLM, (ii) the {\em deployers}, who are fine-tuning the LLM for their specific use case, (iii) the {\em users} of the LLM, who are generating outputs from the LLM directly or from a fine-tuned model, or (iv) the {\em recipients}, who consume the products generated by users.

\smallskip

From the developer point of view, due to the complexity of LLMs, it would be impractical to list all the risks the model generated by an LLM entails within every possible specific usage domain, for example health care. Similarly, ensuring privacy and security in the use of all such derivative systems would not be feasible. This would put an undue burden of compliance to the direct regulation of developers of LLMs. It is therefore argued in \cite{HACK23} that regulation should concentrate on the deployers and users of LLM technology rather than on the LLM developers themselves. In particular, regulation needs to address the capability of LLMs to generate inappropriate and/or unsafe content. Moreover, since certification includes the process by which products are assessed as conforming to standards, all the above considerations will apply with regards to the application of any regulatory standards, concentrating on the ability of the LLM deployers and users to demonstrate compliance, rather than the LLM developers. We note that LLM developers are still accountable under existing regulations, including data protection, non-discrimination and online safety.

\section{Concluding remarks}
\label{sec:conc}

We have reviewed some of the research we are doing at NPL in the area of TAI from a metrology perspective, with emphasis on uncertainty quantification. In Section~\ref{subsec:app1} we reviewed explainable AI, which is a central theme in TAI related to faireness and accountability in AI systems and to metrological traceability. In Section~\ref{subsec:app2} we focused on the application of ML to regression tasks in metrology, presenting a framework that details the requirements that uncertainty quantification for ML should meet; see Figure~\ref{requirements}. In Section~\ref{subsec:app3} we presented a good practice framework for the preparation of training data for marine autonomous navigation systems that use ML; see Figure~\ref{training_framework}. In Section~\ref{sec:cert} we summarised some of the issues pertaining to the certification of AI systems stressing the importance of adherence to the principles of TAI; see \cite{LEVE23}.

\smallskip

The work at NPL focuses both on understanding and elaborating the principles of TAI in the context of metrology, and on the application of these principles to areas of high impact in science and engineering including healthcare, climate, environment, autonomous vehicles and advanced engineering. We believe that there is significant potential of this kind of work to provide high benefits for society.

\section*{Acknowledgments}

Some of the work we have described was supported by the UK Government’s Department for Business, Energy and Industrial Strategy (BEIS) and Department for Science, Innovation and Technology (D-SIT). The work described in Section~\ref{subsec:app3} is based upon an output of a collaboration between NPL and Lloyd’s Register Group Limited. The authors wish to thank Stuart Chalmers (NPL) for reviewing the document and making useful suggestions.

\bibliographystyle{alpha}
\bibliography{tai}

\newcommand{\etalchar}[1]{$^{#1}$}
\begin{thebibliography}{PSGH{\etalchar{+}}21}

\bibitem[ABA{\etalchar{+}}21]{Antoran2021a}
J.~Antoran, U.~Bhatt, T.~Adel, A.~Weller, and J.~Hernandez-Lobato.
\newblock {Getting a CLUE: A method for explaining uncertainty estimates}.
\newblock {\em {International Conference on Learning Representations (ICLR)}},
  2021.

\bibitem[ADD{\etalchar{+}}20]{ARRI20}
A.B. Arrieta, N.~{D{\'\i}az-Rodr{\'\i}guez}, J.~{Del Ser}, A.~Bennetot, et~al.
\newblock Explainable artificial intelligence {(XAI)}: {C}oncepts, taxonomies,
  opportunities and challenges toward responsible {AI}.
\newblock {\em Information Fusion}, 58:82--115, 2020.

\bibitem[AGW18]{Adel2018}
T.~Adel, Z.~Ghahramani, and A.~Weller.
\newblock {Discovering interpretable representations for both deep generative
  and discriminative models}.
\newblock {\em International Conference on Machine Learning (ICML)}, pages
  50--59, 2018.

\bibitem[BCD{\etalchar{+}}12]{bich2012revision}
W.~Bich, M.G. Cox, R.~Dybkaer, C.~Elster, et~al.
\newblock Revision of the `{G}uide to the expression of uncertainty in
  measurement'.
\newblock {\em Metrologia}, 49:702, 2012.

\bibitem[BHA{\etalchar{+}}22]{BOMM22}
R.~Bommasani, D.A. Hudson, E.~Adeli, R.~Altman, et~al.
\newblock On the opportunities and risks of foundation models.
\newblock {\em Machine Learning Archive}, arXiv:2108.07258 [cs.LG], 2022.

\bibitem[BII{\etalchar{+}}08a]{bipm2008}
BIPM, IEC, IFCC, ILAC, ISO, IUPAC, IUPAP, and OIML.
\newblock Evaluation of measurement data -- {G}uide to the expression of
  uncertainty in measurement ({GUM} 1995 with minor corrections).
\newblock {\em Joint Committee for Guides in Metrology (JCGM)}, 100, 2008.

\bibitem[BII{\etalchar{+}}08b]{bipm2008supplement}
BIPM, IEC, IFCC, ILAC, ISO, IUPAC, IUPAP, and OIML.
\newblock Evaluation of measurement data -- {S}upplement 1 to the `{G}uide to
  the expression of uncertainty in measurement'.
\newblock {\em Joint Committee for Guides in Metrology (JCGM)}, 101, 2008.

\bibitem[BII{\etalchar{+}}08c]{JCGM12}
BIPM, IEC, IFCC, ILAC, ISO, IUPAC, IUPAP, and OIML.
\newblock International vocabulary of metrology -- {B}asic and general concepts
  and associated terms (vim).
\newblock {\em Joint Committee for Guides in Metrology (JCGM)}, 2008.
\newblock 3rd edition (2008 version with minor corrections).

\bibitem[Bol19]{Bologna2019a}
G.~Bologna.
\newblock {A simple convolutional neural network with rule extraction}.
\newblock {\em {Applied Sciences}}, page 2411, 2019.

\bibitem[BTTP23]{bilson2023machine}
S.~Bilson, A.~Thompson, D.~Tucker, and J.~Pearce.
\newblock A machine learning approach to automation and uncertainty evaluation
  for self-validating thermocouples.
\newblock submitted for publication, 2023.

\bibitem[CCP18]{wetton_ccpi_2018}
{CCPI} {Europe} signs license agreement with {NPL} for {INSEVA} thermocouple.
\newblock
  \url{https://ccpi-europe.com/2018/05/22/inseva-thermocouple-license-signing/},
  2018.
\newblock Accessed: 15-07-2022.

\bibitem[CDFM20]{CROW20}
S.~Crowder, C.~Delker, E.~Forrest, and N.~Martin.
\newblock {\em Introduction to Statistics in Metrology}.
\newblock Springer Nature, Cham, Switzerland, 2020.

\bibitem[COL72]{COLREGs1972}
{C}onvention on the {I}nternational {R}egulations for {P}reventing {C}ollisions
  at {S}ea ({COLREG}s).
\newblock \url{https://www.imo.org/en/About/Conventions/Pages/COLREG.aspx},
  1972.
\newblock Accessed: 30-06-2023.

\bibitem[FKS{\etalchar{+}}22]{FREE22}
L.~Freeman, J.~Kauffman, D.~Sobien, T.~Cody, and E.~Lanus.
\newblock Best practices for addressing new challenges in testing and
  evaluating artificial intelligence enabled systems.
\newblock {\em AIRC Perspectives}, September 2022.
\newblock 11 pages.

\bibitem[FS11]{forbes2011gum}
A.~B. Forbes and J.~A. Sousa.
\newblock The {GUM}, {B}ayesian inference and the observation and measurement
  equations.
\newblock {\em Measurement}, 44(8):1422--1435, 2011.

\bibitem[GBC17]{GOOD17}
I.~Goodfellow, Y.~Bengio, and A.~Courville.
\newblock {\em Deep Learning}.
\newblock Adaptive Computation and Machine Learning series. MIT Press,
  Cambridge, MA, 2017.

\bibitem[GBY{\etalchar{+}}18]{Gilpin2018a}
L.~Gilpin, D.~Bau, B.~Yuan, A.~Bajwa, M.~Specter, and L.~Kagal.
\newblock {Explaining explanations: An overview of interpretability of machine
  learning}.
\newblock {\em {International Conference on Data Science and Advanced
  Analytics}}, pages 80--89, 2018.

\bibitem[Gil10]{GILL10}
J.~Gillard.
\newblock An overview of linear structural models in errors in variables
  regression.
\newblock {\em REVSTAT - Statistical Journal}, 8:57--80, 2010.

\bibitem[{Goo}23]{GOOG23}
{Google}.
\newblock Meet {Bard}.
\newblock \url{https://bard.google.com}, 2023.

\bibitem[Hai16]{Hailesilassie2016a}
T.~Hailesilassie.
\newblock {Rule extraction algorithm for deep neural networks: A review}.
\newblock {\em Computer Vision and Pattern Recognition Archive},
  arXiv:1610.05267 [cs.CV], 2016.

\bibitem[HEM23]{HACK23}
P.~Hacker, A.~Engel, and M.~Mauer.
\newblock Regulating {ChatGPT} and other large generative {AI} models.
\newblock {\em Computers and Society Archive}, arXiv:2302.02337 [cs.CY], 2023.

\bibitem[HW21]{HULL21}
E.~H\"{u}llermeier and W.~Waegeman.
\newblock Aleatoric and epistemic uncertainty in machine learning: an
  introduction to concepts and methods.
\newblock {\em Machine Learning}, 110:457--506, 2021.

\bibitem[JCCS19]{joskowicz2019inter}
L.~Joskowicz, D.~Cohen, N.~Caplan, and J.~Sosna.
\newblock Inter-observer variability of manual contour delineation of
  structures in {CT}.
\newblock {\em European Radiology}, 29:1391--1399, 2019.

\bibitem[KKL{\etalchar{+}}21]{karniadakis2021physics}
G.~E. Karniadakis, I.~G. Kevrekidis, L.~Lu, P.~Perdikaris, et~al.
\newblock Physics-informed machine learning.
\newblock {\em Nature Reviews Physics}, 3(6):422--440, 2021.

\bibitem[KKSI23]{KUMA23}
Y.~Kumar, A.~Koul, R.~Singla, and M.~F. Ijaz.
\newblock Artificial intelligence in disease diagnosis: a systematic literature
  review, synthesizing framework and future research agenda.
\newblock {\em Journal of Ambient Intelligence and Humanized Computing},
  14:8459--–8486, 2023.

\bibitem[KT21]{khatry2021good}
R.~Khatry and A.~Thompson.
\newblock Good practice in training set preparation for marine navigation
  systems.
\newblock Technical report, National Physical Laboratory, 2021.
\newblock Technical report MS-33.

\bibitem[KURD22]{KAUR22}
D.~Kaur, S.~Uslu, K.~J. Rittichier, and A.~Durresi.
\newblock Trustworthy artificial intelligence: {A} review.
\newblock {\em ACM Computing Surveys}, 55:Article 39, 38 pages, 2022.

\bibitem[KWM{\etalchar{+}}15]{klauenberg2015tutorial}
K.~Klauenberg, G.~W{\"u}bbeler, B.~Mickan, P.~Harris, and C.~Elster.
\newblock A tutorial on {B}ayesian normal linear regression.
\newblock {\em Metrologia}, 52(6):878, 2015.

\bibitem[LAE{\etalchar{+}}16]{liu2016ssd}
W.~Liu, D.~Anguelov, D.~Erhan, C.~Szegedy, et~al.
\newblock Ssd: Single shot multibox detector.
\newblock In {\em Computer Vision--ECCV 2016: 14th European Conference,
  Amsterdam, The Netherlands, October 11--14, 2016, Proceedings, Part I 14},
  pages 21--37, 2016.

\bibitem[Li95]{Li1995a}
P.~Li.
\newblock {\em Neural networks for automatic arc welding}.
\newblock PhD thesis, PhD thesis, 1995.

\bibitem[LQL{\etalchar{+}}23]{LI23}
B.~Li, P.~Qi, B.~Liu, S.~Di, et~al.
\newblock Trustworthy {AI}: {F}rom principles to practices.
\newblock {\em ACM Computing Surveys}, 55:Article 177, 46 pages, 2023.

\bibitem[LW23]{LEVE23}
M.~Levene and J.~Wooldridge.
\newblock Certification of machine learning applications in the context of
  trustworthy {AI} with reference to the standardisation of {AI} systems.
\newblock {NPL Report MS} 45, National Physical Laboratory (NPL), March 2023.
\newblock See \url{https://doi.org/10.47120/npl.MS45}.

\bibitem[LZL{\etalchar{+}}18]{lary2018machine}
D.~J. Lary, G.~K. Zewdie, X.~Liu, D.~Wu, et~al.
\newblock Machine learning applications for earth observation.
\newblock In {\em Earth observation open science and innovation}, volume 165.
  Springer Cham, Switzerland, 2018.

\bibitem[{Mat}23]{MATH23}
{Mathmet European Metrology Network for Mathematics and Statistics}.
\newblock Strategic research agenda.
\newblock
  \url{https://www.euramet.org/european-metrology-networks/mathmet/strategy/strategic-research-agenda},
  2023.
\newblock See Section 3, Strategic Topic – Artificial Intelligence and
  Machine Learning.

\bibitem[MHHS19]{mansour2019automated}
A.~Mansour, A.~Hassan, W.~M. Hussein, and E.~Said.
\newblock Automated vehicle detection in satellite images using deep learning.
\newblock In {\em International Conference on Aerospace Sciences and Aviation
  Technology}, volume~18, pages 1--8. The Military Technical College, 2019.

\bibitem[MWYY20]{MA20}
Y.~Ma, Z.~Wang, H.~Yang, and L.~Yang.
\newblock Artificial intelligence applications in the development of autonomous
  vehicles: A survey.
\newblock {\em IEEE/CAA Journal of Automatica Sinica}, 7:315--329, 2020.

\bibitem[NRR{\etalchar{+}}22]{Niri2022a}
M.~Niri, C.~Reynolds, L.~Ramirez, E.~Kendrick, and J.~Marco.
\newblock {Systematic analysis of the impact of slurry coating on manufacture
  of Li-ion battery electrodes via explainable machine learning}.
\newblock {\em {Energy Storage Materials}}, 51:223--238, 2022.

\bibitem[NYC15]{Nguyen2015a}
A.~Nguyen, J.~Yosinski, and J.~Clune.
\newblock {Deep neural networks are easily fooled: {H}igh confidence
  predictions for unrecognizable images}.
\newblock {\em {IEEE Conference on Computer Vision and Pattern Recognition}},
  pages 427--436, 2015.

\bibitem[{Ope}23]{CHAT23}
{OpenAI}.
\newblock Introducing {ChatGPT}.
\newblock \url{https://openai.com/blog/chatgpt}, 2023.

\bibitem[PA16]{patgiri2016big}
R.~Patgiri and A.~Ahmed.
\newblock Big data: The v's of the game changer paradigm.
\newblock In {\em 2016 IEEE 18th international conference on high performance
  computing and communications; IEEE 14th international conference on smart
  city; IEEE 2nd international conference on data science and systems
  (HPCC/SmartCity/DSS)}, pages 17--24, 2016.

\bibitem[PSGH{\etalchar{+}}21]{Poursabzi-Sangdeh2021a}
F.~Poursabzi-Sangdeh, D.~Goldstein, J.~Hofman, J.~Vaughan, and H.~Wallach.
\newblock {Manipulating and measuring model interpretability}.
\newblock {\em CHI conference on human factors in computing systems}, pages
  1--52, 2021.

\bibitem[PT90]{preston-thomas_international_1990}
H.~Preston-Thomas.
\newblock The {International} {Temperature} {Scale} of 1990 ({ITS}-90).
\newblock {\em Metrologia}, 27(1):3--10, January 1990.
\newblock Publisher: IOP Publishing.

\bibitem[RHC{\etalchar{+}}23]{robinson2023impact}
S.~Robinson, P.~Harris, S-H. Cheong, L.~Wang, et~al.
\newblock Impact of the {COVID}-19 pandemic on levels of deep-ocean acoustic
  noise.
\newblock {\em Scientific Reports}, 13(1):4631, 2023.

\bibitem[RN21]{RUSS21}
S.~Russell and P.~Norvig.
\newblock {\em Artificial Intelligence: A Modern Approach}.
\newblock Pearson Series In Artificial Intelligence. Pearson Education,
  Hoboken, NJ, fourth edition, 2021.

\bibitem[RSG16]{Ribeiro2016a}
M.~Ribeiro, S.~Singh, and C.~Guestrin.
\newblock {Why should {I} trust you? Explaining the predictions of any
  classifier}.
\newblock {\em {ACM SIGKDD International Conference on Knowledge Discovery and
  Data Mining}}, pages 1135--1144, 2016.

\bibitem[Sej19]{SEJN19}
T.J. Sejnowski.
\newblock The unreasonable effectiveness of deep learning in artificial
  intelligence.
\newblock {\em Proceedings of the National Academy of Sciences of the United
  States of America}, 117:30033--30038, 2019.

\bibitem[SNN22]{Sadowski2015a}
L.~Sadowski, M.~Nikoo, and M.~Nikoo.
\newblock {Principal component analysis combined with a self organization
  feature map to determine the pull-off adhesion between concrete layers }.
\newblock {\em {Construction and Building Materials}}, 78:386--396, 2022.

\bibitem[SVZ13]{Simonyan2013a}
K.~Simonyan, A.~Vedaldi, and A.~Zisserman.
\newblock {Deep inside convolutional networks: Visualising image classification
  models and saliency maps}.
\newblock {\em Computer Vision and Pattern Recognition Archive},
  arXiv:1312.6034 [cs.CV], 2013.

\bibitem[Tab23]{TABA23}
E.~Tabassi.
\newblock Artificial intelligence risk management framework ({AI RMF} 1.0).
\newblock Technical report, NIST Trustworthy and Responsible AI, National
  Institute of Standards and Technology, Gaithersburg, MD, January 2023.
\newblock See \url{https://doi.org/10.6028/NIST.AI.100-1}.

\bibitem[TAE{\etalchar{+}}18]{tucker_integrated_2018}
D.~Tucker, A.~Andreu, C.~Elliott, T.~Ford, et~al.
\newblock Integrated self-validating thermocouples with a reference temperature
  up to 1329 °{C}.
\newblock {\em Measurement Science and Technology}, 29:105002, 9 pages, 2018.

\bibitem[TG99]{Taha1999a}
I.~Taha and J.~Ghosh.
\newblock {Symbolic interpretation of artificial neural networks}.
\newblock {\em {IEEE Transactions on Knowledge and Data Engineering}},
  11:448--463, 1999.

\bibitem[Thr95]{Thrun1995a}
S.~Thrun.
\newblock {Extracting rules from artificial neural networks with distributed
  representations}.
\newblock {\em {Advances in Neural Information Processing Systems (NeurIPS)}},
  1995.

\bibitem[TJS{\etalchar{+}}21]{thompson2021uncertainty}
A.~Thompson, K.~Jagan, A.~Sundar, R.~Khatry, et~al.
\newblock Uncertainty evaluation for machine learning.
\newblock Technical report, National Physical Laboratory, 2021.
\newblock Technical report MS-34.

\bibitem[Tor09]{torralba2009many}
A.~Torralba.
\newblock How many pixels make an image?
\newblock {\em Visual neuroscience}, 26(1):123--131, 2009.

\bibitem[TRS{\etalchar{+}}16]{thomas2016dimensionality}
S.~A. Thomas, A.~M. Race, R.~T. Steven, I.~S. Gilmore, and J.~Bunch.
\newblock Dimensionality reduction of mass spectrometry imaging data using
  autoencoders.
\newblock In {\em 2016 IEEE Symposium Series on Computational Intelligence
  (SSCI)}, pages 1--7, 2016.

\bibitem[Var22]{VARS22}
K.~R. Varshney.
\newblock {\em Trustworthy Machine Learning}.
\newblock Independently Published, Chappaqua, NY, 2022.
\newblock See \url{http://www.trustworthymachinelearning.com}.

\bibitem[VOC]{VOC2007}
{PASCAL} {V}isual {O}bject {C}lasses {C}hallenge 2007 ({VOC}2007) {A}nnotation
  {G}uidelines.
\newblock \url{http://host.robots.ox.ac.uk/pascal/VOC/voc2007/guidelines.html}.
\newblock Accessed: 30-06-2023.

\bibitem[YZDP20]{yang2020transfer}
Q.~Yang, Y.~Zhang, W.~Dai, and Sinno~J. Pan.
\newblock {\em Transfer learning}.
\newblock Cambridge University Press, 2020.

\bibitem[ZCAW17]{Zintgraf2017a}
L.~Zintgraf, T.~Cohen, T.~Adel, and M.~Welling.
\newblock {Visualizing deep neural network decisions: Prediction difference
  analysis}.
\newblock {\em {International Conference on Learning Representations (ICLR)}},
  2017.

\bibitem[ZCG{\etalchar{+}}20]{zoph2020learning}
B.~Zoph, E.~D. Cubuk, G.~Ghiasi, T-Y. Lin, et~al.
\newblock Learning data augmentation strategies for object detection.
\newblock In {\em Computer Vision--ECCV 2020: 16th European Conference,
  Glasgow, UK, August 23--28, 2020, Proceedings, Part XXVII 16}, pages
  566--583, 2020.

\bibitem[ZF14]{Zeiler2014a}
M.~Zeiler and R.~Fergus.
\newblock {Visualizing and understanding convolutional networks}.
\newblock {\em {European conference on computer vision}}, pages 818--833, 2014.

\end{thebibliography}

\end{document}